\documentclass[11pt,a4paper]{article}
\usepackage[hyperref]{emnlp-ijcnlp-2019}
\usepackage{times}
\usepackage{latexsym}

\usepackage{url}

\aclfinalcopy 

\usepackage[utf8]{inputenc}
\inputencoding{utf8}
\usepackage{CJKutf8}
\usepackage{multirow}
\usepackage{lipsum}
\usepackage{url}
\usepackage{amsmath}
\usepackage{bm}
\usepackage{wrapfig}
\usepackage{graphicx}
\usepackage[export]{adjustbox}

\title{\texttt{uniblock}: Scoring and Filtering Corpus\\with Unicode Block Information}

\author{
Yingbo Gao \qquad Weiyue Wang \qquad Hermann Ney \\
Human Language Technology and Pattern Recognition Group \\
Computer Science Department\\
RWTH Aachen University \\
D-52056 Aachen, Germany \\
{\tt <surname>@i6.informatik.rwth-aachen.de}
}

\date{}

\begin{document}
\maketitle
\begin{abstract}
The preprocessing pipelines in Natural Language Processing usually involve a step of removing sentences consisted of illegal characters. The definition of illegal characters and the specific removal strategy depend on the task, language, domain, etc, which often lead to tiresome and repetitive scripting of rules. In this paper, we introduce a simple statistical method, \texttt{uniblock}\footnote{The source code is available at \url{https://github.com/ringoreality/uniblock}}, to overcome this problem. For each sentence, \texttt{uniblock} generates a fixed-size feature vector using Unicode block information of the characters. A Gaussian mixture model is then estimated on some clean corpus using variational inference. The learned model can then be used to score sentences and filter corpus. We present experimental results on Sentiment Analysis, Language Modeling and Machine Translation, and show the simplicity and effectiveness of our method.
\end{abstract}

\section{Introduction}

Identification and removal of sentences with illegal characters is a common heuristic in the preprocessing pipelines in Natural Language Processing (NLP). While it has benefits of controlling the vocabulary size and dropping noisy data, it is often a tedious work to come up with appropriate rules and removal strategies when facing with different tasks, languages and domains. The lack of clear definition of what is illegal exacerbates the problem. For example, modern Chinese text may allow characters such as: traditional and simplified Chinese characters, special punctuation marks, full-width characters, emojis, mathematical symbols, Latin characters, currency symbols, scientific notations, etc. As a result, scripting robust rules often requires a considerable amount of time and effort.

In this paper, we introduce a simple statistical method, \texttt{uniblock}, to address the problem. The motivation of our approach is straightforward - good rules may vary greatly depending on the situation, therefore instead of designing rules by hand, we use ``rules" defined by data. We assume that some clean corpus is available, in which all the characters are deemed legal and the character distributions are similar to the test cases. It is possible to learn a probabilistic model, which describes the clean corpus and assigns scores to sentences in another corpus. The scores can further be used for filtering. Since the legalness of characters is in question, Unicode block information is a natural choice for obtaining feature vectors. Note that, by designing alternative feature vectors, one can potentially adapt \texttt{uniblock} to implement other corpus filtering heuristics.

We develop \texttt{uniblock} mainly for Machine Translation (MT). It can also be easily applied to other NLP tasks. We present experimental results on Sentiment Analysis (SA), Language Modeling (LM) and MT, and show the simplicity and effectiveness of our method.

\section{Related Work}

Raw data in NLP is often noisy. \citet{khayrallah2018impact} categorize five common noise sources in parallel corpora and count only about 23\% of the sentences in the raw 2016 ParaCrawl corpus\footnote{\url{https://paracrawl.eu/index.html}} to be ``Okay". Although illegal characters is not listed as a separate noise source, misuse of characters and shifting of character distributions may result in a sentence being classified into one of the five noise sources. In previous work, a supervised model using bag-of-words translation features is developed to classify clean and noisy data \citep{xu2017zipporah, khayrallah2018jhu}. In contrast, our model, which is trained in an unsupervised manner, tackles the illegal character problem explicitly.

\citet{koehn2018findings} describe a shared task on parallel corpus filtering. While participating systems focus on addressing both monolingual fluency and bilingual adequacy, character-level filtering is common to all submissions. \citet{junczys2018dual} applies a language identification model to implicitly remove sentences with illegal characters. \citet{rossenbach2018rwth} keep sentences with more than three words, with each word having at least one character from the predefined alphabet of the language. \citet{lu2018alibaba} remove characters outside of a predefined alphabet. \citet{ash2018speechmatics} count most frequent characters, set a cutoff around eighty for each language, and remove sentences with illegal characters. \citet{erdmann2018coverage} get rid of lines containing characters from the Unicode general category of ``other". \citet{papavassiliou2018ilsp} simply consider Latin Unicode characters to be legal.

Unicode is the de facto standard for encoding characters from various languages, domains and sources \cite{Unicode:19}. It uses ``blocks" to group characters with similar origins or functions. The current version 12.0 defines 300 blocks, including Basic Latin, Latin-1 Supplement, CJK (Chinese, Japanese and Korean) Symbols and Punctuation, etc. To identify the legalness of characters, the Unicode block information provides meaningful discriminative signals.

The Gaussian Mixture Model (GMM) is a classic algorithm that assumes data is generated from a mixture of finite number of Gaussian distributions, whose parameters are typcially estimated with the Expectation–Maximization (EM) algorithm. An extension to the EM algorithm is variational inference, which has the advantage of automatically choosing the number of components. \citet{bishop2006pattern} gives a comprehensive introduction to the topic. We use the implementation of variational Bayesian estimation of Gaussian mixtures from \texttt{scikit-learn} \cite{scikit-learn}.

\section{Methodology}

\subsection{Feature Vectors}

In order to assign meaningful scores to sentences and eventually filter out those who contain illegal characters, we first need to design appropriate features. We believe, the Unicode block information has a good property that the blocks are grouped by origin and function. For instance, CJK Symbols and Punctuation has a dedicated Unicode block in range U+3000...U+303F. Specifically, we count the appearances of characters in each of the 300 blocks. This will result in feature vectors $c_1, c_2, ..., c_B$ with a fixed length of $B=300$. If we further normalize them by the total number of characters in each sentence, probability distributions over the 300 blocks can be obtained. We observe that there are only few blocks whose counts are non-zero in natural language texts. This calls for dimensionality reduction methods. Empirically, we drop the zero-count dimensions directly during training and assign conceptually low scores\footnote{zeros in our experiments} when a non-zero count is seen during scoring. That is, we use normalized feature vectors $e=e_1, e_2, ..., e_{B'}$, where $1, 2, ..., B'$ are dimensions in $B$ whose original counts are non-zero, for training.

\subsection{Bayesian Gaussian Mixture Model}

Although many corpora are noisy, it is not appropriate to deem all sentences in them ``dirty". While generating synthetic noisy data is always an option, it is unrealistic to cover all types of noises. Compared to the difficulty to obtain negatively labelled data, the development set is often available and can be deemed ``clean" with high confidence. Therefore, we take the development set as training data for our scoring system and treat the problem as a clustering task rather than a classification task.

We assume that the training feature vectors $e$ are generated by a mixture of Gaussian distributions.
\begin{equation} \label{eq:BGMM}
\begin{aligned}
p(e) &= \sum_{k=1}^{K} \pi_k \mathcal{N}(e|\mu_k,\Sigma_k)\\
\pi_k &\sim \mathcal{DP}(\alpha)\\
\mu_k &\sim \mathcal{N}(\mu_0, \sqrt{\frac{1}{\tau}})\\
\Sigma_k &\sim \mathcal{W}_{B'} (V, n) 
\end{aligned}
\end{equation}
In the equation above, $k$ is a running index in the number of mixtures $K$, $\pi_k$ is the mixture weight for the $k$-th Gaussian, $\mu_k$ is a $B'$-dimensional vector parametrizing the $k$-th mean, $\Sigma_k$ is the $k$-th $B' \times B'$ covariance matrix. We further impose priors on the model parameters: $\pi_k$ follows a Dirichlet Process ($\mathcal{DP}$), which is parametrized by concetration prior $\alpha$; $\mu_k$ follows a Normal distribution ($\mathcal{N}$), which is parametrized by the $B'$-dimensional mean prior $\mu_0$ and the precision prior $\tau$; $\Sigma_k$ follows a Wishart distribution ($\mathcal{W}$) in $B'$, which is parametrized by the covariance prior $V$ and degree of freedom prior $n$. We estimate the model parameters using the EM algorithm. Note that operating in $B'$ dimensions leads to significantly better model convergence than training in $B$ dimensions.

\subsection{Scoring and Filtering}

Once the model is trained till convergence, it is possible to use it to assign scores to unseen feature vectors. We directly use the weighted log probabilities as scores. Compared to the sentences used during \texttt{uniblock} training, higher scored sentences have more similar Unicode block count distributions, or fewer illegal characters. Regarding how to remove noisy sentences, we implement two straightforward ideas: absolute thresholding and relative thresholding. The former removes sentences with scores lower than a given threshold while the later removes a given percentage of low-scored sentences. We also add support for filtering parallel data. A simple method is to combine scores across parallel sentences with a specified reduction method (minimum, maximum, average), and then apply thresholds. Alternatively, one can use the lowest score seen during training as an absolute threshold for each language and filter parallel data in a ``one bad, remove all" manner.

\section{Experiments}

We conduct experiments on three NLP tasks: SA, LM and MT. As shown in related work, many different heuristics are used for the purpose of illegal character filtering, and there is no generally accepted ``golden rule". Therefore, we focus on comparing ``no filtering" versus ``filtering", to examine how our method performs.

\subsection{Sentiment Analysis}

We conduct SA experiments on the Stanford Twitter Sentiment (STS)
corpus \cite{go2009twitter} using fastText \cite{joulin2017bag}. Given 1.6M automatically annotated tweets, the task is to predict postive or negative sentiments on 498 test tweets. For the baseline, we use minimal preprocessing to handle casing, weblinks, usernames, hashtags and spacing in the raw training tweets, without explicit character-level filtering. For \texttt{uniblock}, we additionally train a GMM on the tweets in the test set\footnote{No develepment set is available for STS, this is why we trained the GMM on the test set. For LM and MT experiments, \texttt{uniblock} is trained on the development sets.} and use the minimum score as an absolute threshold to filter the training corpus. In total, about 0.9\% training tweets are filtered out. We observe that only one Unicode block exists in the test set, which means all sentences with characters in other blocks are assigned conceptually low scores and removed. In this particular case, our general method reduces to a simple filtering rule similar to that of \citet{papavassiliou2018ilsp}. As shown in Table \ref{tab:SA}, our method improves the test accuracy over the baseline by 0.6\%.

\begin{table}[h!]
\centering
\begin{tabular}{cc}
\hline
& test accuracy [\%] \\ \hline
baseline & 81.6 \\
\texttt{uniblock} & \textbf{82.2} \\ \hline
\end{tabular}
\caption{Test accuracies on STS.}
\label{tab:SA}
\end{table}

\subsection{Language Modeling}\label{sec:lm}

In MT pipelines, language models are often used for corpus filtering, fusion techniques and $N$-best reranking. Therefore, we perform LM experiments on MT corpora. Specifically, we take monolingual data from parallel corpora of WMT19 \citep{barrault-EtAl:2019:WMT} Chinese-English (zh-en) and Russian-English (ru-en) as training data. We use newsdev2017 for both sides of zh-en and newstest2016 for both sides of ru-en as development data. We concatenate newstest2017 and newstest2018 as test data for all four sides. The sizes of the corpora (WMT) are summarized in Table \ref{tab:numsents}.
\begin{table}[h!]
\centering
\begin{tabular}{ccc}
\hline
 & zh-en & ru-en \\ \hline
train & 26M & 25M \\
valid & 2K & 3K \\
test & 6K & 6K \\ \hline
\end{tabular}
\caption{Number of sentence pairs in WMT.}
\label{tab:numsents}
\end{table}

\begin{table}[h!]
\centering
\begin{tabular}{ccccc}
\hline
\multirow{2}{*}{} & \multicolumn{2}{c}{zh-en} & \multicolumn{2}{c}{ru-en} \\
data & zh & en & ru & en \\ \hline
100\% & 40.9 & 135.0 & 181.0 & \textbf{145.7} \\
90\% & \textbf{36.9} & 133.5 & \textbf{176.7} & 147.6 \\
80\% & 37.1 & 132.9 & 180.6 & 149.3\\
70\% & 38.7 & \textbf{131.4} & 178.6 & 151.1\\ \hline
\end{tabular}
\caption{Test perplexities on WMT.}
\label{tab:testppl}
\end{table}

\begin{table*}[h!]
\centering
\begin{tabular}{crr}
\hline
side & score & parallel text \\ \hline
zh & -48523.31 & \begin{CJK*}{UTF8}{gkai}\underline{リリースノート}\end{CJK*} \\
en & 53.82 & Open Directory - Unix \\ \hline
zh & 67.71 & \begin{CJK*}{UTF8}{gkai}健康、整洁、卫生和清洁技术组织\end{CJK*} \\
en & -158.57 & Salubrit\underline{é}, propret\underline{é}, hygi\underline{è}ne et techniques d’assainissement \\ \hline
zh & 0.00 & \begin{CJK*}{UTF8}{gkai}从系统目录\underline{→}位置\underline{→}传送我的位置\end{CJK*}... \\
en & 0.00 & \begin{CJK*}{UTF8}{gkai}From System Menu \underline{→} Location \underline{→} Send My Location\end{CJK*}... \\ \hline
zh & 36.01 & \begin{CJK*}{UTF8}{gkai}在25\underline{℃}下测定了样品的总交换容量和平衡等温线.\end{CJK*} \\
en & 40.65 & \begin{CJK*}{UTF8}{gkai}Cation exchange capacity and equilibria isotherm at 25 \underline{℃} were determined. \end{CJK*} \\ \hline
zh & 68.63 & \begin{CJK*}{UTF8}{gkai}财务政策及发展小组\end{CJK*}  \\
en & 53.82 & Financial Policy and Development Unit \\ \hline
\end{tabular}
\caption{\label{tab:scores}Several text samples from the zh-en parallel corpus with \texttt{uniblock} scores. We find empirically: sentences with only foreign characters are scored very low; sentences with some foreign characters are scored low; sentences with characters in unseen blocks are scored 0; sentences with only few foreign characters are scored high; sentences with no illegal characters are scored very high. Foreign (or illegal) characters are underlined.}
\end{table*}

For each of the four LM subtasks, we train \texttt{uniblock} on the corresponding development set and filter out 10\%, 20\% and 30\% of the raw training corpus. We train 2-layer LSTM language models with hidden dimensions being 1024. For the zh subtask, we train character-level language models and include all seen characters in the vocabulary. For the other 3 subtasks, top 100k frequent words are used as the vocabulary and the rest of the words are replaced with unknown tokens. All LM experiments are conducted using the RETURNN framework \cite{doetsch2017returnn, zeyer2018:returnn}. As shown in Table \ref{tab:testppl}, we see consistent improvements in the \texttt{uniblock}-filtered setups of zh, ru and en under zh-en. For en under ru-en, the baseline system outperforms \texttt{uniblock}-filtered ones.

\subsection{Machine Translation}

We train MT systems to show the simplicity and effectiveness of our method in a more challenging task. We use the same parallel corpora as in Section \ref{sec:lm} and train systems in four directions: zh-en, en-zh, ru-en and en-ru. The raw training corpora are filtered with \texttt{uniblock} as before. Before examining the performance of the translation systems, we first manually examine some filtered text samples. Five parallel text samples with \texttt{uniblock} scores are shown in Table \ref{tab:scores}. We observe that the trained GMM works properly and indeed assigns low scores to unreasonable or ``dirty'' sentences. We believe that \texttt{uniblock} is superior to other rule-based methods because the relative quality of the sentences is modeled. For example, a hard rule may eliminate a complete sentence pair encountering even one \begin{CJK*}{UTF8}{gkai}℃\end{CJK*} symbol, while \texttt{uniblock} only gives a small penalization due to the fact that one occurrence of the symbol \begin{CJK*}{UTF8}{gkai}℃\end{CJK*} does not alter the count-based feature vector $e$ much.

\begin{table}[h!]
\centering
\begin{tabular}{ccccc}
\hline
 & \multicolumn{2}{c}{zh-en} & \multicolumn{2}{c}{en-zh} \\
data & test17 & test18 & test17 & test18 \\ \hline
100\% & 25.0 & 24.5 & 30.1 & 33.0 \\
90\% & \textbf{25.2} & \textbf{25.6} & \textbf{30.9} & 33.1 \\
80\% & 24.3 & 25.3 & 30.3 & \textbf{33.2} \\
70\% & 24.3 & 24.8 & 30.2 & 33.0 \\ \hline
\\ \hline
 & \multicolumn{2}{c}{ru-en} & \multicolumn{2}{c}{en-ru} \\
data & test17 & test18 & test17 & test18 \\ \hline
100\% & 32.9 & 28.3 & 26.9 & 23.3 \\
90\% & \textbf{33.5} & \textbf{29.3} & \textbf{27.8} & \textbf{23.9} \\
80\% & 33.3 & 28.9 & 27.2 & 23.6 \\
70\% & 32.9 & 28.1 & 26.6 & 23.2 \\ \hline
\end{tabular}
\caption{Case-sensitive \textsc{Bleu}(\%) scores on four different translation tasks.}
\label{tab:mt}
\end{table}

For all MT systems, we perform minimum preprocessing to handle the whitespaces and casing, and use SentencePiece \cite{kudo2018sentencepiece} to obtain the subword units. All models adopt the Transformer \cite{vaswani2017attention} architecture and use the exact same hyperparameters as the base model. Trainings are done with the Sockeye \cite{hieber2017sockeye} toolkit and share the same optimization hyperparameters. No synthetic data is used and no ensembling is done during decoding. The only difference across the models is the training corpus - the baseline model uses the full parallel corpus, while the ones filtered by \texttt{uniblock} use a subset of the full corpus. We take the checkpoint with the lowest perplexity on the validation set and report case-sensitive \textsc{Bleu}(\%) \cite{papineni2002bleu} scores on newstest2017 and newstest2018 using the sacre\textsc{Bleu} \cite{post-2018-call} software. The translation qualities of the systems are shown in Table \ref{tab:mt}. As can be seen, without altering the architecture or the optimization process, only applying \texttt{uniblock} as a corpus filtering step leads to consistent improvements over the baseline systems for all four directions, up to 1.1 \textsc{Bleu}(\%).

Linguistically, the zh-en and ru-en language pairs are rather distant. For more closely-related language pairs such as French and English, our method should also perform well. As shown in Table \ref{tab:scores}, a Chinese sentence could be corrupted with Japanese characters and an English sentence could be corrupted with French characters. For both cases, our method is able to discriminate the illegal characters and assign low scores to these sentences. Note that, our method would fail, if ``clean" and ``dirty" sentences share one exact same block. Because then after normalization all the feature vectors will be essentially the same and therefore indistinguishable. In this case, one should carefully design other features so that the GMM is able to assign meaningful scores. However, when ``clean" and ``dirty" sentences share the same set of blocks, our method still works fine because after normalization the empirical distributions are probably different.

\section{Conclusion}

In this work, we develop a simple and effective statistical method called \texttt{uniblock}. Using Unicode block information as feature vectors, a GMM is estimated with variational inference on some clean corpus, which can then be used to score and filter corpus. We release our implementation which supports parallel corpus filtering and different thresholding. Our experiments show concrete and consistent improvements in SA, LM and MT.

We believe that the method can be extended and improved by examining other dimensionality reduction methods and alternatives to GMM, and by introducing other heuristics into the feature vector, such as sentence length and punctuation marks\footnote{Common punctuation marks lie in the same Unicode block range U+0000..U+007F as the English alphabet and are currently not separated.}.

\section*{Acknowledgements}

\begin{center}
    \includegraphics[width=0.2\textwidth, valign=m]{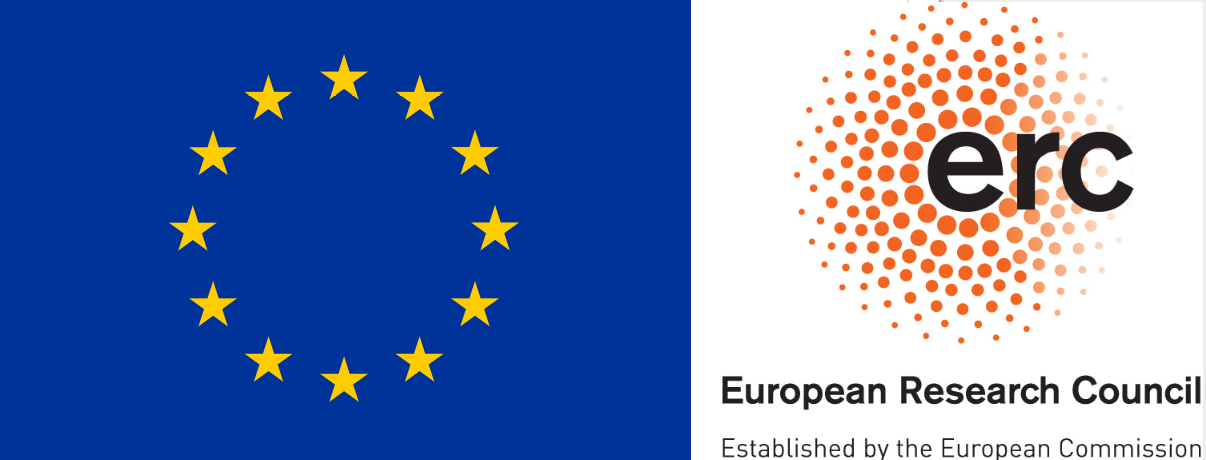}
    \hspace{3mm}
    \includegraphics[width=0.2\textwidth, valign=m]{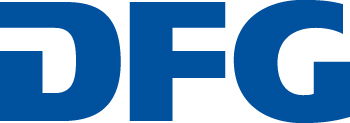}
\end{center}

This work has received funding from the European Research Council (ERC) (under the European Union's Horizon 2020 research and innovation programme, grant agreement No 694537, project "SEQCLAS") and the Deutsche Forschungsgemeinschaft (DFG; grant agreement NE 572/8-1, project "CoreTec"). The GPU computing cluster was supported by DFG (Deutsche Forschungsgemeinschaft) under grant INST 222/1168-1 FUGG.

\bibliography{emnlp-ijcnlp-2019}
\bibliographystyle{acl_natbib}

\end{document}